\documentclass[letterpaper, 10 pt, conference]{ieeeconf}  

\IEEEoverridecommandlockouts

\usepackage{amsmath}
\usepackage{cite}
\usepackage{subcaption}
\usepackage{tabularx}
\usepackage{array}
\usepackage{booktabs}
\usepackage{multirow}
\usepackage{multicol}
\usepackage{graphicx}
\usepackage{caption}
\title{\LARGE \bf
Learning Rapid Turning, Aerial Reorientation, and Balancing using Manipulator as a Tail
}
\author{Insung Yang\textsuperscript{1}, and  Jemin Hwangbo\textsuperscript{1}
\thanks{$^{1}$ Robot and Artificial Intelligence Lab in the Department of Mechanical Engineering, Korea Advanced Institute of Science and Technology, Daejeon, Republic of Korea
        {\tt\small \{ycube13, jhwangbo\} @kaist.ac.kr}}%
}
\begin{document}
\maketitle
\thispagestyle{empty}
\pagestyle{empty}

\begin{abstract}
In this research, we investigated the innovative use of a manipulator as a tail in quadruped robots to augment their physical capabilities. Previous studies have primarily focused on enhancing various abilities by attaching robotic tails that function solely as tails on quadruped robots. While these tails improve the performance of the robots, they come with several disadvantages, such as increased overall weight and higher costs. To mitigate these limitations, we propose the use of a 6-DoF manipulator as a tail, allowing it to serve both as a tail and as a manipulator. To control this highly complex robot, we developed a controller based on reinforcement learning for the robot equipped with the manipulator. Our experimental results demonstrate that robots equipped with a manipulator outperform those without a manipulator in tasks such as rapid turning, aerial reorientation, and balancing. These results indicate that the manipulator can improve the agility and stability of quadruped robots, similar to a tail, in addition to its manipulation capabilities.

\end{abstract}

\section{INTRODUCTION}
The tails of animals play crucial roles in enhancing their physical abilities, particularly in quadrupeds. For instance, cheetahs use their tails to change direction during high-speed chases, allowing them to catch prey more effectively \cite{wilson2013locomotion}. Additionally, squirrels and lizards utilize their tails to adjust body orientation in mid-air, ensuring safe landings \cite{jusufi2010righting, jusufi2008active}. Tails are also essential for balancing, swimming, hopping, courtship, and defense \cite{o2014kangaroo}. These versatile tails have garnered the attention of robotic engineers and have been established as solutions for addressing various challenges in the field of robotics.

Previous works have attached tails to quadruped robots to improve their stability and speed during turning and running \cite{patel2013rapid, patel2014rapid, kohut2013precise, casarez2018steering}. Some studies suggest that tails can aid the robot's attitude control in aerial regions \cite{fukushima2021inertial, chang2011lizard, tang2023towards, johnson2012tail}. There have also been attempts to improve the overall stability of robots by incorporating tails \cite{briggs2012tails, norby2021enabling}. These studies demonstrate that tails can enhance robotic performance in various tasks. However, adding a tail increases the complexity of the robot, leading to issues such as higher costs and increased weight. Consequently, the use of tails in practical applications may not always be efficient.

\begin{figure}
\centering
\includegraphics[width=0.6\linewidth]{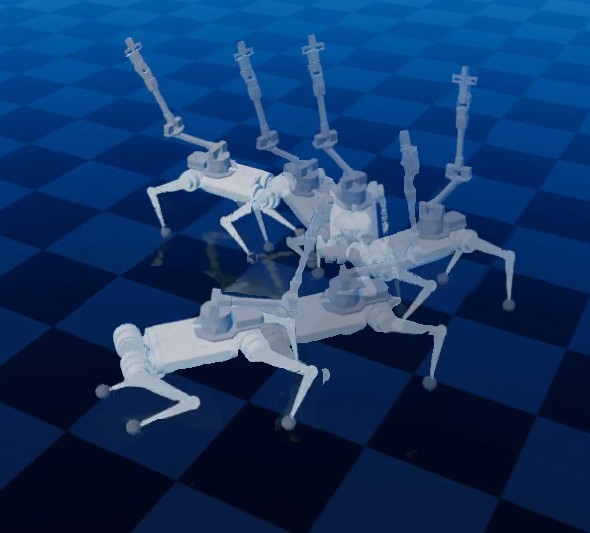}
\caption{A quadruped robot, Mini Cheetah, equipped with a WidowX250S 6-DoF manipulator, is executing a rapid $135^{\circ}$ turn while running at 4.5 m/s. It learns to utilize the manipulator as a tail for sharp turns via reinforcement learning.}
\label{fig
}
\end{figure}
To alleviate these drawbacks and create a highly efficient tail capable of executing various tasks, we employ a manipulator as a tail rather than designing a tail solely for that purpose. Huang et al. \cite{huang2023more} are also exploring the use of a manipulator as a tail. By employing a manipulator as a tail, it can simultaneously perform the roles of a gripper and a tail, thereby achieving high efficiency. In recent years, numerous studies have explored the use of manipulators in quadruped robots \cite{ferrolho2023roloma,fu2023deep,sleiman2021unified,zimmermann2021go}. The majority of these studies have focused on manipulators performing manipulation tasks. However, we have identified the potential for manipulators when utilized as tails.

Among various control methods, reinforcement learning (RL) has recently emerged as one of the most popular approaches for legged robots \cite{hwangbo2019learning,lee2020learning,miki2022learning}. By using a deep reinforcement learning-based controller, many previous studies have successfully enabled high-complexity robots to execute complex behaviors \cite{ma2023learning, rudin2021cat}. Inspired by these studies, we employ deep reinforcement learning (DRL) to develop a controller.

Beyond creating a robust controller, the primary consideration when incorporating a tail on a robot is whether its presence meaningfully enhances the robot's overall performance. Our experiments demonstrate that a robot equipped with a manipulator outperforms one without a manipulator in tasks such as rapid turning, aerial reorientation, and balancing. Specifically, when a robot running at 4.5 m/s executed a $135^{\circ}$ turn, the distance pushed out by centrifugal force was reduced by two-thirds compared to a robot without a manipulator. Additionally, a robot with a manipulator was able to land safely from initial angles of $90^{\circ}$ to $120^{\circ}$ at heights of 1.5 to 2.25 meters, whereas a robot without a manipulator failed to land under the same conditions. Finally, when subjected to external forces, a robot with a manipulator exhibited a higher survival rate compared to one without a manipulator.

Overall, this paper demonstrates that a manipulator can be an appropriate choice as a quadruped robot's tail, enhancing the capabilities of a quadruped robot. Our main contributions are as follows.
\begin{enumerate}
\item We propose the utilization of a manipulator that functions as a tail.
\item We outline a method for controlling a robot equipped with a tail using reinforcement learning.
\item We demonstrate that a robot equipped with a manipulator exhibits improved performance in tasks such as rapid turning, aerial reorientation, and balancing.
\end{enumerate}

\begin{figure*}
    \centering
    \includegraphics[width=0.9\textwidth]{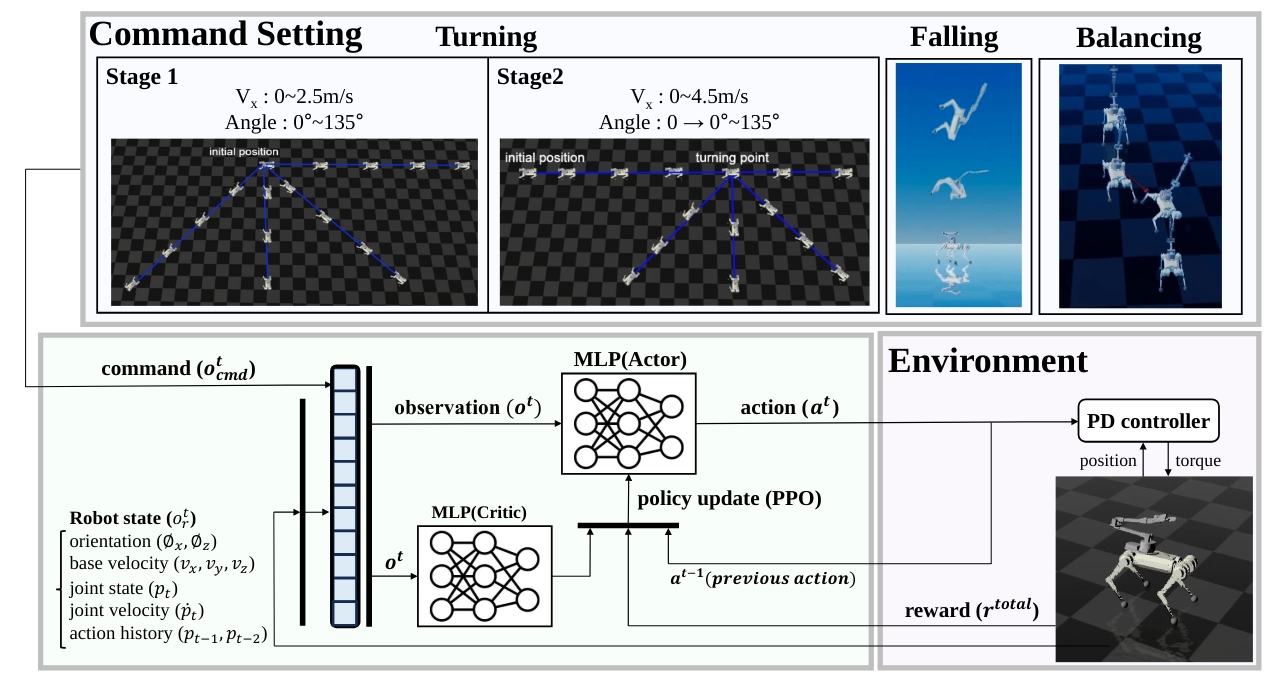}
    \caption{In the proposed reinforcement learning pipeline, two neural networks, namely the Actor and the Critic, are utilized. The Actor network generates the actions, while the Critic network computes the value function, which is essential for updating the Actor network. Both networks are structured as Multilayer Perceptron (MLP) and receive observations as inputs. These observations encompass two types: command, which is specified by the task, and robot state, which provides information about the robot's current state. The actions generated by the Actor are conveyed to the environment through a Proportional-Derivative (PD) controller, and the resulting reward, based on the action's effectiveness, is used to update the Actor network. This updating process employs the Proximal Policy Optimization (PPO) algorithm\cite{schulman2017proximal}, refining the policy governed by the Actor network.}
    \label{fig:method}
\end{figure*}
\section{METHOD}
\subsection{Overview}
Our goal is to develop a controller for a manipulator-mounted quadruped robot that performs well on various tasks, including rapid turning, aerial reorientation for safe landing, and balancing against external force. We utilize deep reinforcement learning to train a neural network policy. An overview of this deep reinforcement learning process is illustrated in Figure~\ref{fig:method}. In our study, we employ Proximal Policy Optimization (PPO), a deep reinforcement learning algorithm, to update the policy and utilize two distinct neural network architectures. The first network, termed the "actor," maps observations to actions and is structured as a Multi-Layer Perceptron (MLP) with hidden layers sized [512, 256, 128]. The second network, known as the "critic," evaluates the current state using an MLP with hidden layers sized [512, 256, 128]. Both networks are essential for the algorithm's decision-making process and performance evaluation, facilitating the implementation of PPO in complex environments.

In our simulation, we deploy this controller on the Minicheetah robot mounted with a WidowX250S manipulator. We use RAISIM \cite{hwangbo2018per} as the simulation environment.
\subsection{Base}
For three different tasks, we employ slightly different learning algorithms. However, the overarching methodology remains largely consistent. This section outlines the general algorithm constant for all tasks, and the subsequent sections will provide a detailed description of the specific adaptations made for each task.
\subsubsection{Observation}
As depicted in Figure~\ref{fig:method}, the observation is divided into two distinct parts: the robot state and the command. The robot state comprises the joint state, base state, and joint history. We provide the joint velocity ($\dot{p}_{t}$) and joint position ($p_{t}$) for the joint state. For the base state, we include the base angular velocity ($w_{x,y,z}$), base linear velocity ($v_{x,y,z}$), and base orientation ($\phi_{x,z}$). Additionally, we provide the previous joint position history for the two preceding time steps ($p_{t-1}$, $p_{t-2}$). This type of observation remain constant across all tasks. 

Unlike the robot state, the observation of commands($o^{t}_{\text{cmd}}$) varies according to the task. For the task of rapid turning, we provide the command velocity and command yaw: $o^{t}_{\text{cmd}} = (V_x, V_y, \phi_{\text{cmd}})$. For the balancing task, we use the same command structure. In contrast, for the task of aerial reorientation, we do not provide any command observations. Since the goal is to land safely when falling in the air, no particular command is required. Without any command, the task can be sufficiently completed by termination.

\begin{table}[t]
\caption{General Constraint Reward}
\label{table:reward_function1}
\centering
\begin{tabular}{|c|c|c|c|}
\hline
\textbf{Reward} & \textbf{Expression} & \multicolumn{2}{|c|}{\textbf{Reward Coefficient}}\\
\hline
Joint Position & \(r_p=k_p\left\|p_t-p_{\text {nominal }}\right\|^2 \) & $k_p$&-4.0\\
\hline
Joint Velocity & \( r_{\dot{p}}=k_{\dot{p}}\left\|\dot{p}_t\right\|^2 \) & $k_{\dot{p}}$&-0.005\\
\hline
Torque & \( r_\tau=k_\tau\|\tau\|^2  \) & $k_\tau$&-0.002\\
\hline
smoothness & \( r_{s}=k_{s}\left\|p_t^{d e s}-q_{t-1}^{d e s}\right\|^2  \) & $k_{s}$&-4.0\\
\hline
\end{tabular}
\end{table}

\subsubsection{Action}
Instead of using the non-scaled output of the actor network, we scale the output by the technique used by Ji et al. \cite{ji2022concurrent}. Here, the action space is scaled by a predefined factor, termed the action factor, and adds to the nominal joint position as follows: $q^{des}_{t}=q^{nominal} + \sigma_{a} \cdot a_{t}$, where $q^{des}_{t}$ is the desired position which would be transmitted to the PD controller ($K_{p}: 17\text{N}\cdot\text{m}\cdot\text{rad}^{-1}$, $K_{p}: 0.4\text{N}\cdot\text{m}\cdot\text{rad}^{-1}$), $q^{nominal}$ is the nominal joint configuration, $\sigma_{a}$ is the action factor, $a_{t}$ is the policy output in the neural network. We use 0.3 for action factor($\sigma_{a}$). In the early iterations, this action scaling largely dominates the action, preventing it from moving excessively away from the initial state. However, as the iterations progress, this dominance diminishes.

\subsubsection{Reward}
We utilize two types of rewards in our system: the objective reward $r^{\text{pos}}$, which grants a positive reward when the desired goal is achieved, and the constraint reward $r^{\text{neg}}$, which imposes a negative reward for unsafe or unfeasible actions. We adopt the total reward formula used by Ji et al. \cite{ji2022concurrent}, where the total reward is calculated as $r^{\text{total}}=r^{\text{pos}} \cdot \text{exp} (reward\, factor\cdot r^{\text{neg}})$. We use 0.02 for $reward\,factor$.

Given that three distinct tasks have different objectives, objective reward $r^{\text{pos}}$ varies for each task. For the constraint reward $r^{\text{neg}}$, it can be divided into two types: the general constraint reward $r^\text{neg}_{gen}$, and the task-specific constraint reward $r^\text{neg}_{task}$. The General constraint reward $r^\text{neg}_{gen}$=$r_p$+$r_{\dot{p}}$+$r_\tau$+$r_{s}$ is illustrated in Table ~\ref{table:reward_function1}, and the task-specific constraint reward will be illustrated in the following sections. In conclusion, the total reward function can be rewritten as follows.

\begin{equation}
    r^{\text{total}}=r^\text{pos}_{task} \cdot \text{exp} (0.02\cdot (r^\text{neg}_{gen}+r^\text{neg}_{task}))
\end{equation}

\subsection{Rapid Turning}
Designing a controller capable of executing rapid turns during high-speed running presents significant challenges. Frequently, the robot becomes trapped in a local minimum, where it remains stationary before executing the turn command and only moves after the command. This phenomenon occurs because the robot encounters substantial difficulty in performing rapid turns while in motion, leading it to abandon the attempt and execute the turn from a stationary position. Although various methods can successfully overcome this problem, such as gradually increasing the turning angle and velocity through curriculum learning, we have found that segmenting the training into two distinct stages is the most effective and stable approach for learning rapid turning. In the first stage, the robot learns to turn from a stationary position. In the second stage, it learns to execute rapid turns while running.

\subsubsection{Stage 1 (Yaw Adjustment in Standstill)}
The concept of Stage 1 is illustrated at Figure~\ref{fig:method}. We train the robot to follow the turn command ranging from $0^{\circ}$ to $135^{\circ}$ in a standstill position. Additionally, the robot learns to follow the velocity command. Stage 1 continues until the iteration count reaches 2000.

To enhance the stability of the learning process, we incrementally increase the command velocity using a curriculum learning approach, which has been shown to improve the stability of learning\cite{ji2022concurrent}. 
\begin{equation}
    V^{cmd}_{x}=1.0 + \frac{1.5}{1.0 + e^{-0.008 \cdot (iteration - 500)}}
\end{equation}

\subsubsection{Stage2 (Rapid turning during running)}
Stage2 is the main stage of rapid turning. In stage 2, the robot is trained to rotate in a fully moving state. The concept is shown at Figure~\ref{fig:method}.

We also use curriculum learning for command velocity in this stage. However, we discover that a straightforward curriculum learning approach by iteration, such as that used in Stage 1, is insufficient. This is mainly because Stage 2 presents greater instability in the learning process compared to Stage 1, due to the requirement for rapid rotation at high speeds. To improve stability, we implement a reward-based curriculum learning strategy\cite{margolis2024rapid}. Instead of merely updating the command velocity based on iteration count alone, we adjust the command velocity if the total reward exceeds a predefined threshold value. 
\begin{equation}
    V^{cmd}_{x}=1.77 + \frac{2.73}{1.0 + e^{-0.01 \cdot (reward\,step - 100)}}
\end{equation} 
In this experiment, the predefined threshold value is set to 4.75. This means that $reward\,step$ is increased by 1 if the total average reward of the iteration exceeds 4.75.

It is essential to determine the proper timing for executing turn commands during the learning phase. If the commands are issued randomly from the beginning of the learning phase, the likelihood of successful learning is reduced because of the continuous variation in the robot's foot position and posture during its run. One method to address this issue is to progressively extend the range of commands using a curriculum learning approach. We update the range of commands if the total reward exceeds a predefined threshold value, similar to the Stage 2 command velocity process.
\begin{equation}
    Command\,Range=1+min(300,reward\,step)
\end{equation}
In this experiment, the value of $reward\,step$ is the same as the $reward\,step$ used in the velocity command. However, it is possible to set different values.
\subsubsection{Reward}
\begin{table}[t]
\caption{Rapid Turning task Reward(Stage2)}
\label{table:reward_function2}
\centering
\begin{tabular}{|c|c|}
\hline
\textbf{Reward} & \textbf{Expression}\\
\hline
Linear Velocity & \(r_v=k_v \exp(-(V^{cmd}_{x}-V_{x})^2-(V^{cmd}_{y}-V_{y})^2) \)\\
\hline
Yaw & \( r_\phi=k_\phi \exp(-2.5angle(\phi^{cmd}_{x},\phi_{x})) \)\\
\hline
Angle Velocity & \( r_w=k_w(3-12\exp(-0.5w_z))  \)\\
\hline
\multicolumn{2}{|l|}{\;Foot Airtime} \\
\hline
\multicolumn{2}{|c|}{$r_{air,i} =
\begin{cases} 
    k_{\text{min}} \cdot \max(T_{s,i}, T_{a,i}, 0.2) & \text{if } T_{\text{max}, i} < 0.25 \\
    0 & \text{otherwise}
\end{cases}$} \\
\hline
Foot Clearance& \( r_{c l, i}=k_{c l}\left(foot\,p_{z,i}-foot\,p_{z}^{thres}\right)^2  \)\\
\hline
Base stability& \( r_{\text {base }}=k_{base}V_z^2  \)\\
\hline
Orientation& \( r_{ori}=k_{ori}(angle(\phi^{world}_{z},\phi^{body}_{z}))^2  \)\\
\hline
Arm position& \(r_{\text{arm}}=k_{\text{arm}}\left\|p^{\text{arm}}_t-p^{\text{arm}}_{\text {nominal }}\right\|^2 \)\\
\hline
\end{tabular}
\end{table}
\begin{table}[t]
\begin{tabular}{|l|c|c|l|c|c|}
\hline
\textbf{Coefficient\;\;\,} & \textbf{\;\,run\;\,}&\textbf{\;\,turn\;\,}&\textbf{Coefficient\;\;\,} & \textbf{\;\,run\;\,}&\textbf{\;\,turn\;\,}\\
\hline
$k_v$&2.0&2.0&$k_\phi$&4.5&4.5\\
\hline
$k_w$&0.0&3.0&$k_{air}$&0.5&0.5\\
\hline
$k_{cl}$&-50.0&-50.0&$k_{base}$&-10.0&-10.0\\
\hline
$k_{ori}$&-100.0&0.0&$k_{arm}$&-15.0&0.0\\
\hline
\end{tabular}
\end{table}
\begin{table}[t]
\caption{Aerial Reorientation and Safe Landing task Reward}
\label{table:reward_function3}
\centering
\begin{tabular}{|c|c|}
\hline
\textbf{\;\;\;\;\;\;\;\;\;Reward\;\;\;\;\;\;\;\;\;} & \textbf{\;\;\;\;\;\;\;\;\;Expression\;\;\;\;\;\;\;\;\;}\\
\hline
Orientation & \( r_{\textbf{ori}}=k_{\textbf{ori}} \exp(-2.5(angle(\phi^{cmd}_{x},\phi_{x}))^2) \)\\
\hline
Linear Velocity & \( r_v=k_v \exp(-5.0(V^{cmd}_{x,y}-V_{x,y})^2) \)\\
\hline
Base height & \( r_h=k_h \exp(-10.0\,|p^{nominal}_z-p_z| \)\\
\hline
Foot Clearance& \( r_{c l, i}=k_{c l}\left(foot\,p_{z,i}-foot\,p_{z}^{thres}\right)^2  \)\\
\hline
Arm position& \(r_{\text{arm}}=k_{\text{arm}}\left\|p^{\text{arm}}_t-p^{\text{arm}}_{\text {nominal }}\right\|^2 \)\\
\hline
\end{tabular}
\end{table}
\begin{table}[t]
\begin{tabular}{|l|c|c|l|c|c|}
\hline
\textbf{Coefficient\;\;} & \textbf{\;air\;}&\textbf{\;ground\;}&\textbf{Coefficient\;\;} & \textbf{\;air\;}&\textbf{\;ground\;}\\
\hline
$k_{\textbf{ori}}$&5.0&5.0&$k_v$&0.0&2.5\\
\hline
$k_h$&0.0&5.0&$k_{cl}$&0.0&-100.0\\
\hline
$k_{arm}$&0.0&-150.0\\
\cline{1-3}
\end{tabular}
\end{table}
Table ~\ref{table:reward_function2} illustrates the reward details for the rapid turning task in Stage 2. We divide each iteration into two distinct sections: one for running forward and the other for turning. Due to the distinct objectives of these two sections, we employ different sets of reward coefficients. Notably, in the turning section, a reward for angular velocity is added, while rewards for body orientation and arm position are omitted. The key factor enabling the robot to turn rapidly is the angular velocity reward ($r_w$). This reward provides a high incentive if the robot's angular velocity is high during the turning phase. It is possible to increase the rotational speed of the robot without using the angular velocity reward. For example, accelerating the robot's lateral speed or increasing the rate of change in the robot's direction can increase the turning speed. However, the angular velocity reward provides the best performance in terms of both learning speed and turning performance. An important consideration when setting up an angular velocity reward is to set the angular velocity reward coefficient ($k_w$) higher than the other objective reward coefficients($k_v,k_\phi,k_{air}$). Without this adjustment, achieving high-speed turning becomes nearly impossible.

Unlike Stage 2, in Stage 1 we solely utilize the reward associated with the running section, without dividing the iteration into two sections.

The objective task reward can be written as $r^\text{pos}_{task}=r_v+r_\phi+r_w+r_{air}$, and the constraint task reward would be $r^\text{neg}_{task}=r_{cl}+r_{base}+r_{ori}+r_{arm}$.
\subsection{Aerial Reorientation and Safe Landing}
Unlike the rapid turning task, we found that the aerial reorientation task does not require curriculum learning or stage-wise learning. The key factors for these tasks are termination as constraint and the inertia of the tail.

\subsubsection{Inertia of tail}
In the task, we utilize not only the WidowX250S, which is used in the other tasks, but also the ViperX300S, which is heavier and longer. We observed that, due to the limited inertia of the tail, the WidowX250S does not perform well in aerial reorientation. Consequently, we employ a robot arm with higher inertia. The WidowX250S has a total length of 1300 mm and a weight of 2.35 kg, whereas the ViperX300S has a total length of 1500 mm and a weight of 3.6 kg.

\subsubsection{Termination}
Reinforcement learning that relies solely on a reward function can have limitations. In this task, the robot receives a substantial reward if it lands safely and incurs a significant penalty if the landing fails. Due to this extreme setup, the robot might prematurely forfeit the reward in the air, as forfeiting the reward is more advantageous than failing to land. To address this issue, precisely adjusting the reward function can be a solution, but implementing a termination constraint could be more effective.

Several studies have attempted to train robots using constrainted reinforcement learning \cite{kim2024not,gangapurwala2020guided,lee2023evaluation,chane2024cat}. Notably, Chane-Sane et al. \cite{chane2024cat} propose the CaT(Constraint as Termination) algorithm which utilize terminator as the constraint in RL method. In our study, both the reward function and termination constraints are used, and no other complex algorithms are necessary. We simply terminate the iteration with a high negative reward when the robot violates certain rules. This simple algorithm, combined with reward function, successfully enables the robot to execute feasible motions. The following rules are used for termination in this experiment:
\[
\begin{cases} 
    \text{Collision with ground except for feet} & \\
    smoothness: r_s > -2.0 \cdot k_s & \\
    torque: r_\tau > -180.0 \cdot k_\tau & \\
    joint\,position: r_p > -5.0 \cdot k_p
\end{cases}
\]
\subsubsection{Reward}
This task consists of two stages: the aerial reorientation section, where the robot rotates its body to align its base orientation with the z-axis of the world frame, and the self-righting section, where the robot recovers its body to the nominal position after landing. We divide these sections according to body height: the aerial reorientation section corresponds to the air region ($p_z>0.4$), and the self-righting section corresponds to the ground region($p_z<0.4$). Due to the different objectives of these two stages, the rewards are divided accordingly, as detailed in Table ~\ref{table:reward_function3}. In particular, in the air, only orientation rewards and general constraint rewards are needed. However, on the ground, where self-righting is required, various additional rewards are included.

The objective task reward can be written as $r^\text{pos}_{task}=r_{ori}+r_v+r_w+r_h$, and the constraint task reward would be $r^\text{neg}_{task}=r_{cl}+r_{arm}$.

\subsection{Balancing}
\begin{figure}
    \centering
    \includegraphics[width=\linewidth]{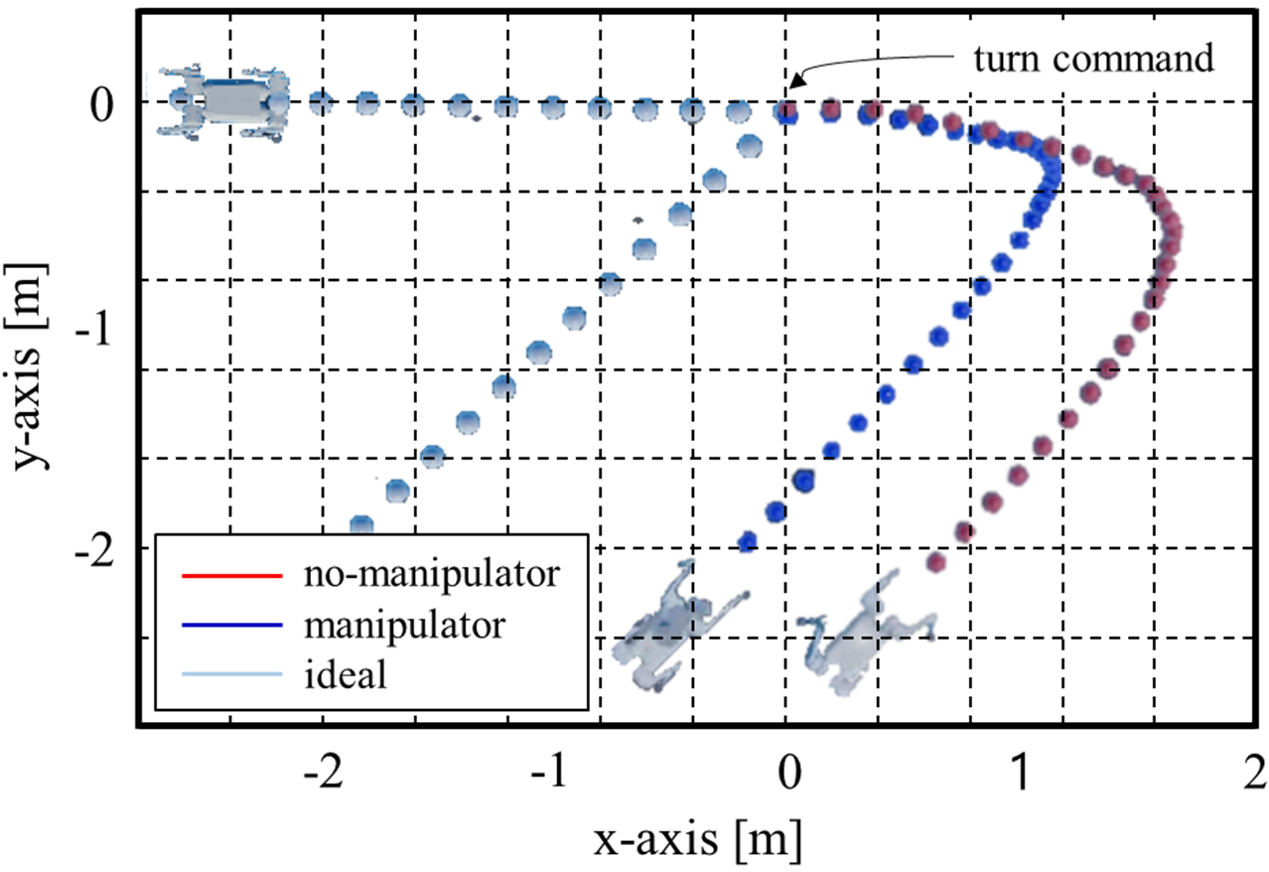}
    \caption{These graphs depict the comparison of trajectories for a robot with and without a manipulator executing a $135^\circ$ turn while running at 4.5 m/s. A dot is plotted every 0.05 seconds. The blue dots indicate the trajectory of the robot equipped with a manipulator following the turn command, while the red dots represent the trajectory of the robot without a manipulator after the turn command. Additionally, the light blue dots denote the ideal trajectory, which assumes an instantaneous turn upon command.}
    \label{fig:result1}
\end{figure}
For the balancing part, the robot is trained to endure random external impulses at random timings and in random directions while walking at speeds ranging from 0.5 m/s to 3 m/s. The range of impulses is $50N\cdot s \sim 100N\cdot s$. During training, we use the same reward setup as in the running section of the Rapid Turning task.

\section{RESULT}

\subsection{Rapid turning}
We compared the turning performance of robots with and without a manipulator in simulations using the RAISIM environment. In the simulation tests, we set the forward velocity of the robots between $3,\text{m/s}$ and $4.5,\text{m/s}$. We then issued commands for the robots to execute turns up to $135^{\circ}$ while moving forward. The quality of the turns was assessed based on three criteria: the speed of the turn, the sharpness of the turn, and the displacement during turning.

Our findings indicate that the turning speed was nearly identical regardless of the presence of the manipulator. However, there was a notable difference in the sharpness of the turns. As depicted in Figure~\ref{fig:result1}, the robot equipped with a manipulator made sharper turns compared to the robot without a manipulator. Furthermore, the trajectory of the robot with a manipulator was closer to the ideal trajectory than that of the robot without a manipulator. This means that the displacement during rotation was shorter when a manipulator was equipped. Specifically, during a $135^{\circ}$ turn, the robot without a manipulator was pushed out by up to 1.74 meters, while the robot with a manipulator was pushed out to a maximum of only 1.2 meters.

\begin{figure}  
    \centering
    \begin{subfigure}[t]{0.49\linewidth}
        \includegraphics[width=\textwidth]{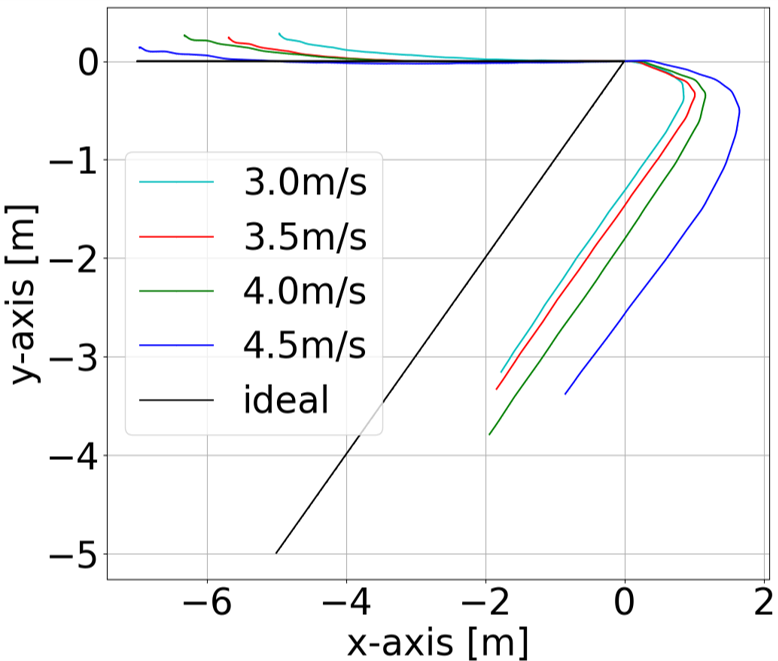}
        \caption{without manipulator}
        \label{fig:result2_a}
    \end{subfigure}
    \begin{subfigure}[t]{0.49\linewidth}
        \includegraphics[width=\textwidth]{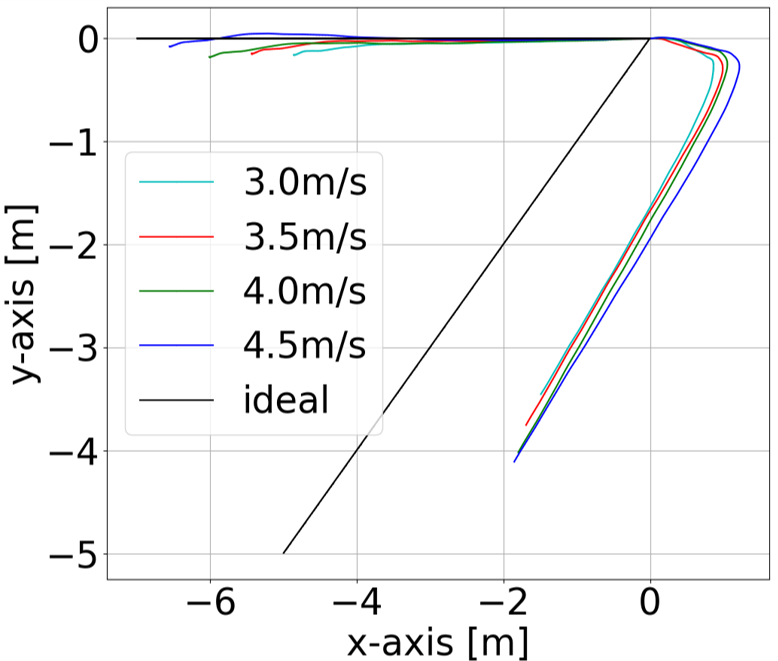}
        \caption{with manipulator}
        \label{fig:result2_b}
    \end{subfigure}
        \caption{These graphs depict the trajectory results of a robot executing a $135^{\circ}$ turn while running at speeds of 3.0 m/s, 3.5 m/s, 4.0 m/s, and 4.5 m/s.}
        \label{fig:result2}
\end{figure}

Figure~\ref{fig:result2} shows the difference in trajectory with respect to velocity. In Figure~\ref{fig:result2_a}, the trajectory of the robot without a manipulator varies significantly with speed. As speed increases, the deviation from the ideal trajectory becomes more pronounced. In contrast, the trajectory of the robot with a manipulator, shown in Figure~\ref{fig:result2_b}, remains relatively consistent.
\begin{figure*}  
    \centering
    \begin{subfigure}[t]{0.32\textwidth}
        \includegraphics[width=\textwidth]{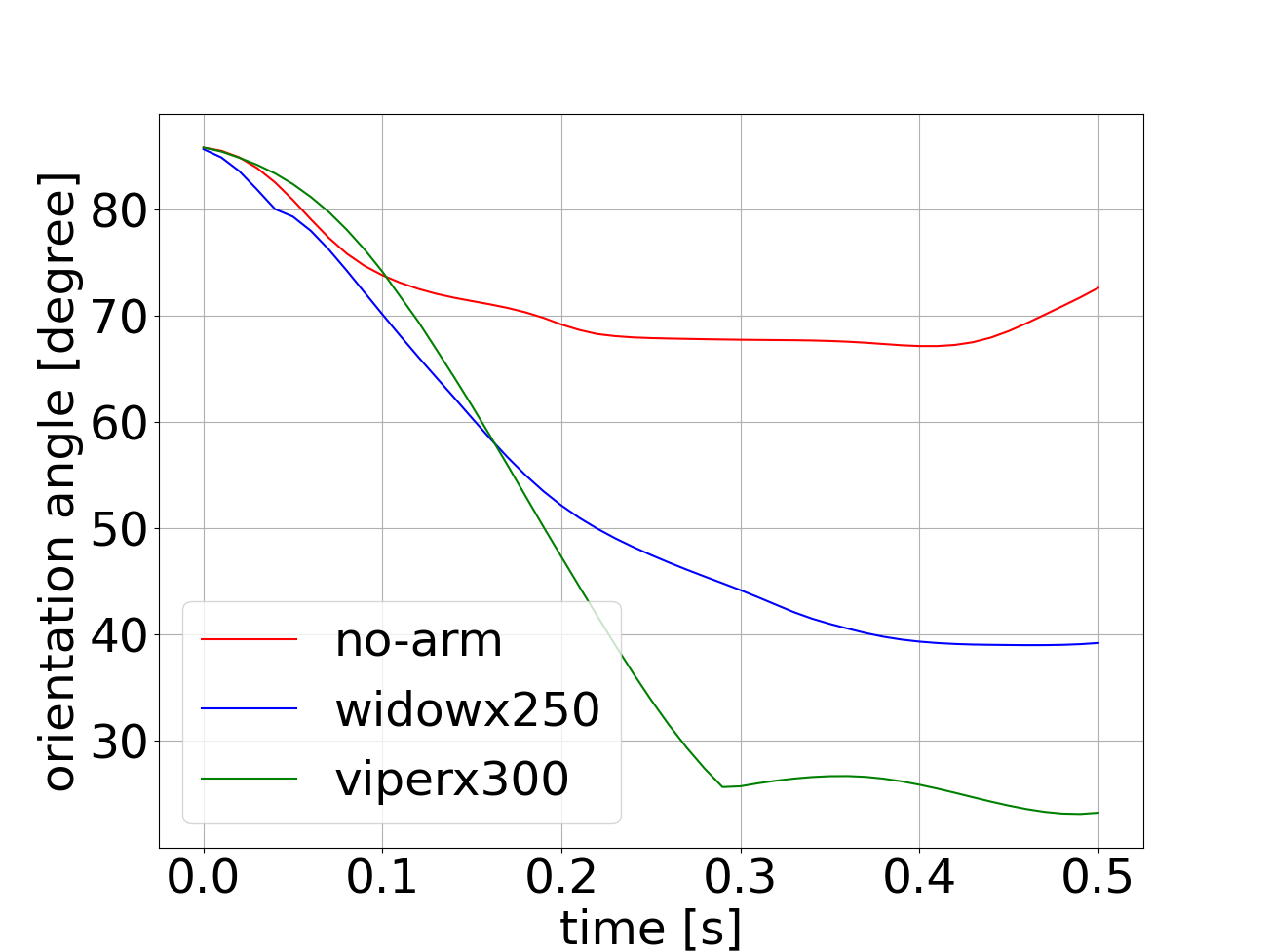}
        \caption{drop at $90^{\circ}$}
        \label{fig:result3_a}
    \end{subfigure}
    \begin{subfigure}[t]{0.32\textwidth}
        \includegraphics[width=\textwidth]{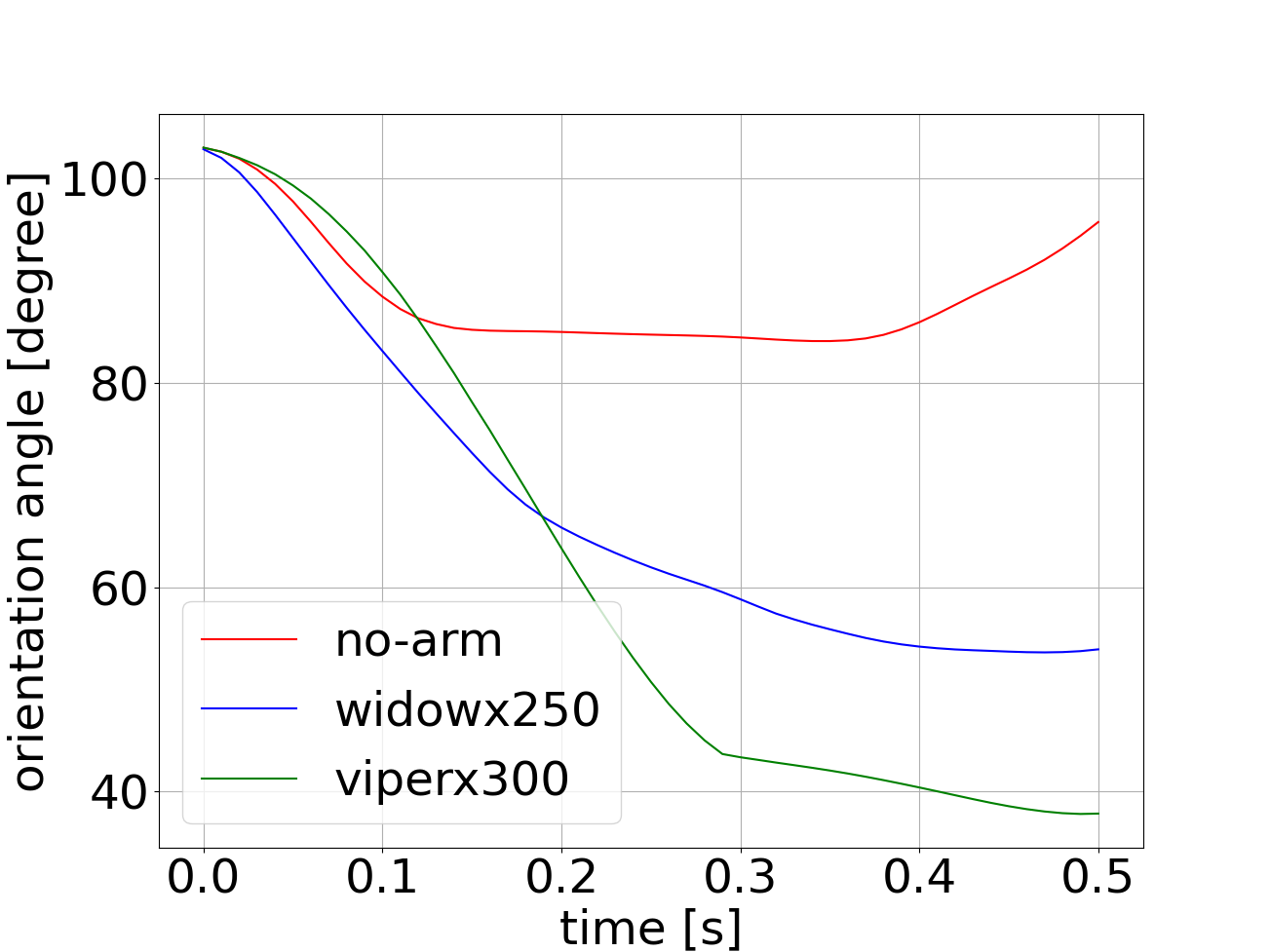}
        \caption{drop at $105^{\circ}$}
        \label{fig:result3_b}
    \end{subfigure}
    \begin{subfigure}[t]{0.32\textwidth}
        \includegraphics[width=\textwidth]{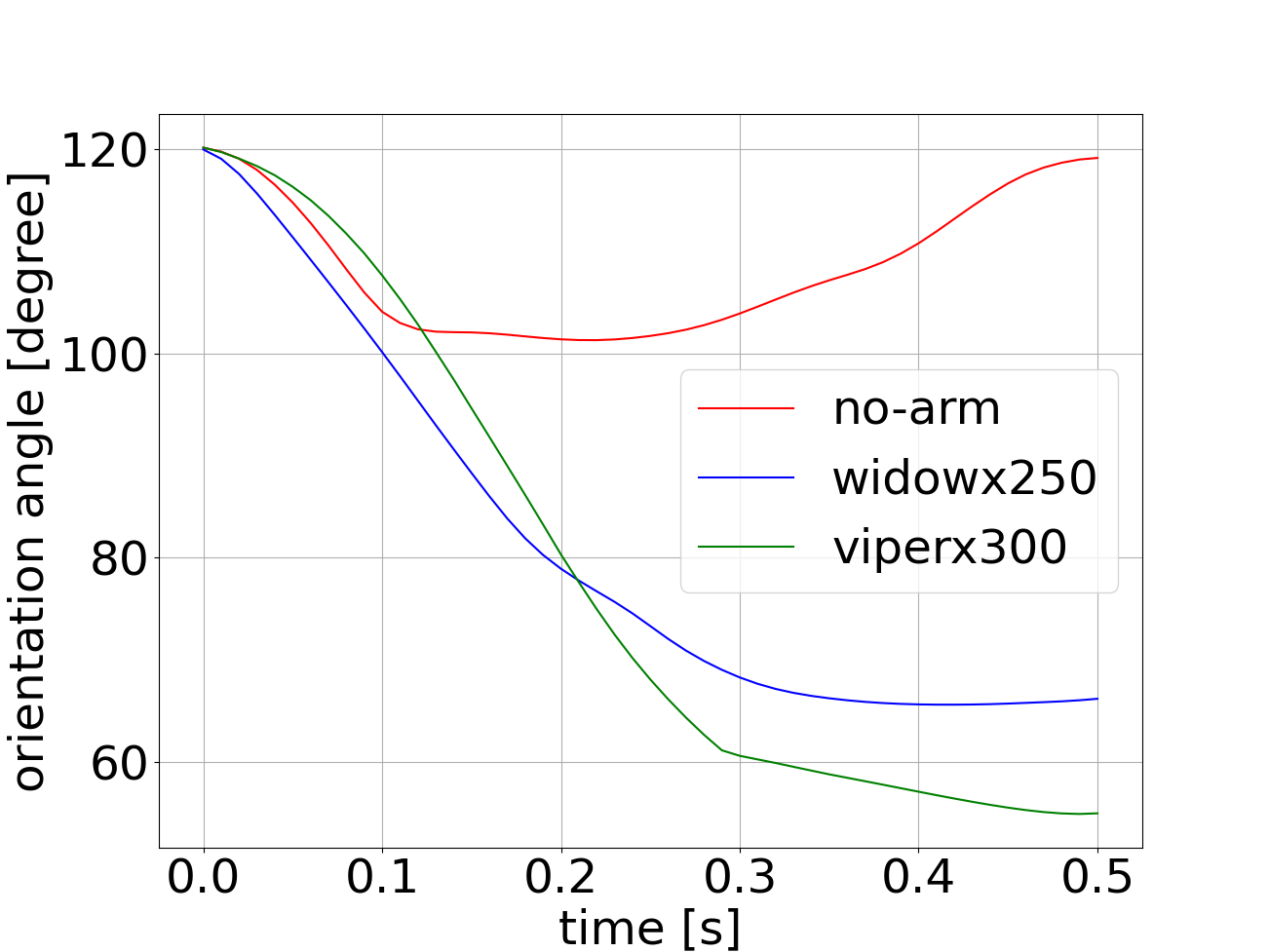}
        \caption{drop at $120^{\circ}$}
        \label{fig:result3_c}
    \end{subfigure}
        \caption{These graph show the angle between the body z-axis and the world z-axis over time from $0,\text{s}$ to $0.5,\text{s}$. The red line shows the result of the robot without a manipulator, blue line shows the robot with a WidowX250S, and the green line shows the robot with a ViperX300S.}
        \label{fig:result3}
\end{figure*}
\subsection{Aerial Reorientation and Safe Landing}
Unlike the previous task, we used three different robots to evaluate performance: Minicheetah without a manipulator, Minicheetah equipped with a WidowX250S, and Minicheetah equipped with a ViperX300S. Details of these three robots are provided in the Method section. In the simulation, we dropped these robots from heights ranging from 1.5m to 2.25m and angles ranging from $90^\circ$ to $120^\circ$.

We found that the robot without a manipulator failed in all these scenarios. Similarly, the robot equipped with a WidowX250S failed in all scenarios. However, the robot equipped with a ViperX300S succeeded in all situations.

Figure~\ref{fig:result3} illustrates the orientation shift during aerial reorientation. Regardless of the initial conditions, the robot without a manipulator achieved a maximum rotation of $15^\circ$, the robot equipped with the WidowX250S reached up to $50^\circ$, and the robot equipped with the ViperX300S attained up to $65^\circ$. The graph for the robot equipped with the WidowX250S shows that its rotation speed decreases after 0.2 seconds. In contrast, the robot equipped with the ViperX300S maintains a consistent rotation speed while reorienting and culminates in a constant orientation angle phase after 0.3 seconds, which remains steady while falling.

\subsection{Balancing}
\begin{figure}  
    \centering
    \includegraphics[width=\linewidth]{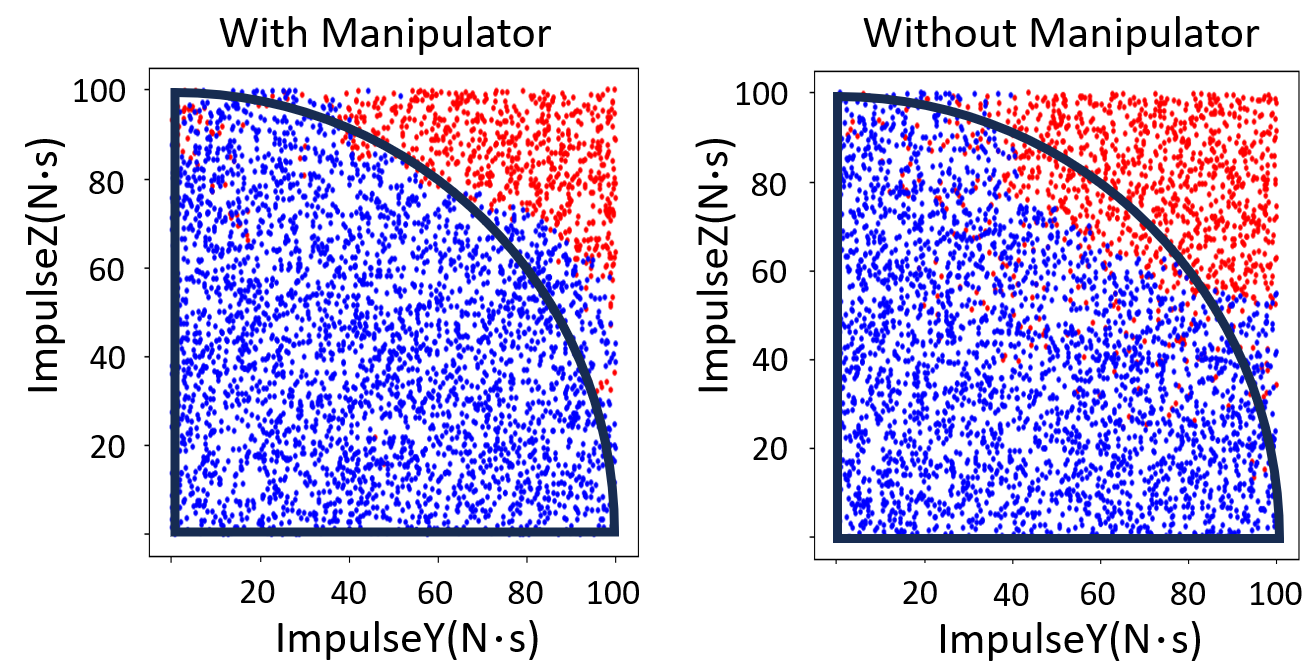}
    \caption{These graphs illustrate the survival outcomes given specific impulse combinations. A red dot indicates that the robot failed to survive at that impulse combination, whereas a blue dot indicates that it survived. The black line delineates the training distribution region. The area within the black line represents the impulse distribution during training.}
    \label{fig:result4}
\end{figure}
\begin{figure*}
    \centering
    \includegraphics[width=1.0\textwidth]{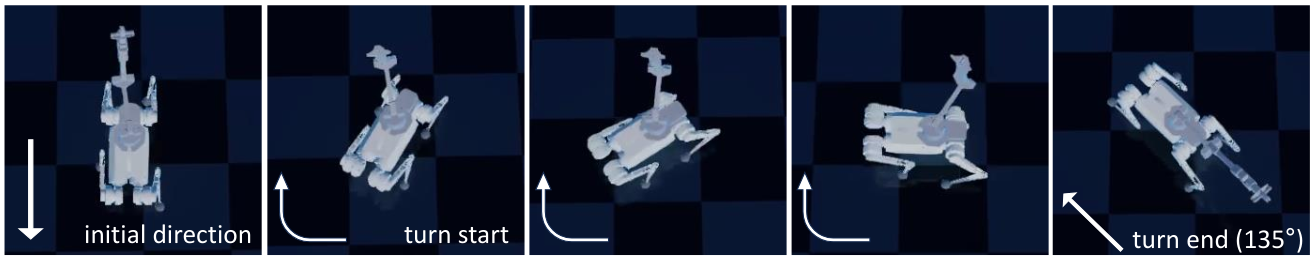}
    \caption{This figure illustrates the movement of the manipulator as the robot executes a $135^{\circ}$ turn. When the robot moves forward, the manipulator is positioned centrally within the robot. However, during the turn, the robot experiences centrifugal force. To counteract this force, the manipulator moves in the opposite direction. Upon completing the turn, the manipulator returns to its original central position.}
    \label{fig:dis1}
\end{figure*}
In simulation, we randomly applied impulses to the robot while walking at 1 m/s, with forces in the x, y and z-axis directions, totaling up to 100 N·s. The x-axis aligns with the robot’s forward direction, the y-axis corresponds to the robot’s lateral direction, and the z-axis aligns with gravity. We only display the results for the impulses on the y- and z-axis because the data show that the robot can withstand forces in the x-axis direction well, both with and without manipulator.

The results are illustrated in Figure~\ref{fig:result4}. The survival rate of the robot equipped with a manipulator is $81.5\%$, while the survival rate of the robot without a manipulator is $71.5\%$. Notably, in almost all areas within the black line, where the robot has already experienced the learning session, the robot with a manipulator successfully survived. In contrast, the robot without a manipulator has relatively often failed in these regions.

\section{Discussion}
\subsection{Rapid Turning}
In the Results section, we evaluate the turning performance of the quadruped robot equipped with a manipulator based on three criteria: the speed of the turn, the sharpness of the turn, and the displacement during turning. Among these criteria, the quadruped robot with a manipulator demonstrates superior performance in terms of sharpness and displacement during the turn. This improved performance is attributed to the robot with the manipulator being able to withstand greater centrifugal force during the turn. Figure~\ref{fig:result2} strongly supports this observation. As velocity increases, the deviation between the ideal trajectory and the trajectory of the robot without the manipulator becomes more pronounced. Given that the magnitude of the centrifugal force increases proportionally with the square of the velocity, this result is expected. However, the trajectory of the robot equipped with the manipulator remains almost constant regardless of the velocity, indicating that the quadruped robot can better endure centrifugal forces with the assistance of the manipulator.

The mechanism by which the quadruped robot utilizes its manipulator to withstand centrifugal force is illustrated in Figure~\ref{fig:dis1}. It shows the movement of the quadruped robot with the manipulator when executing a $135^{\circ}$ turn while running at a speed of 4.5 m/s. During turning, the manipulator is positioned on the side opposite to the direction of the centrifugal force. By positioning itself in this manner, the manipulator generates a counter-torque that opposes the torque produced by the centrifugal force. Consequently, this counter-torque enables the robot to execute sharper turns.

\subsection{Aerial reorientation and safe landing}
\begin{figure}
    \centering
    \includegraphics[width=\linewidth]{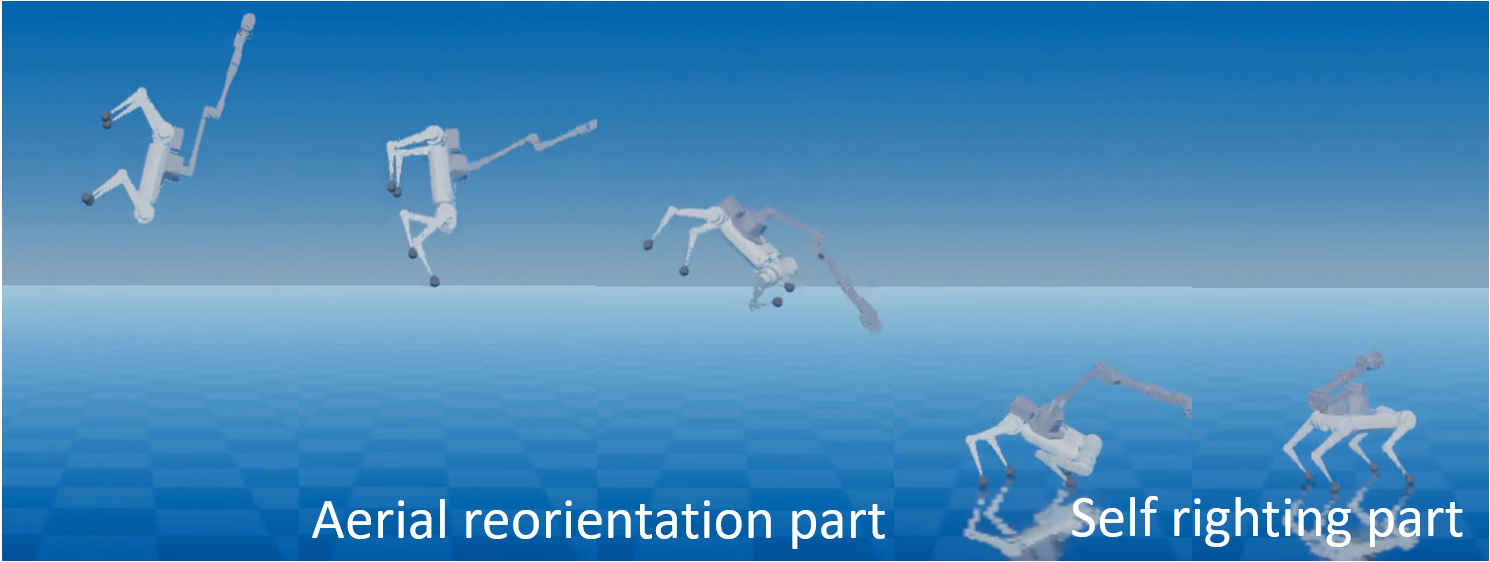}
    \caption{This figure illustrates the motion of the robot as it falls from a height of 1.5 meters with a body angle of $105^{\circ}$. The task comprises two phases: the aerial reorientation phase, where the robot reorients its body to align with the world z-axis, and the self-righting phase, where the robot returns its body to its nominal position.}
    \label{fig:dis2}
\end{figure}

As illustrated in Figure~\ref{fig:result3}, the rotation speed and the total rotation angle vary depending on the type of robot used. Robots equipped with manipulators rotate more rapidly and achieve a larger total rotation angle than a robot without a manipulator, although the extent of these differences varies depending on the specific manipulator used. The disparity in robot performance according to the presence of a manipulator is attributed to the additional inertia provided by the manipulator. Figure~\ref{fig:dis2} depicts the overall process by which the robot utilizes its manipulator when falling upside-down. During the aerial reorientation phase shown in the figure, the robot rotates its manipulator in the opposite direction of the robot base rotation. This movement generates counter momentum, allowing the robot base to reorient its body in the air.

The rotational performance in the air also varies depending on the type of arm used. This is because the counter momentum generated by the manipulator relies on its inertia. Therefore, the robot equipped with the ViperX300S, which has a higher inertia than the WidowX250S, rotates faster and achieves a larger total rotation angle compared to the one equipped with the WidowX250S.

In Figure~\ref{fig:result3}, after approximately 0.3 seconds, there is a region where the robot's orientation remains constant for the robot equipped with the ViperX300S. This region represents the phase where the robot adjusts its legs into a posture advantageous for a safe landing. This phase can also be seen in Figure~\ref{fig:dis2}, where, in the middle of the figure, the robot finishes reorienting its body and adjusts its legs. We found that securing sufficient time to adjust its legs is crucial for a safe landing. For example, the robot equipped with WidowX250S fails to establish this region due to its inadequate speed of rotation in the air, resulting in unsuccessful landing attempts.

\subsection{Balancing}
The experimental results show that the quadruped robot has a higher survival rate when external forces are applied. Therefore, engineers can expect an enhancement in the robot's stability if the manipulator is installed on the quadruped robot. However, when the impulse value is less than $60N\cdot s$, no substantial differences are observed. Consequently, installing a manipulator solely for balance purposes would not be efficient, as situations involving an impulse greater than $60N\cdot s$ are relatively infrequent.

\section{CONCLUSIONS}
In this study, we propose the integration of a 6-DoF manipulator as a multifunctional tail for quadruped robots to address the challenges posed by the complexity of traditional tails. Through simulation, we demonstrated that mounting a manipulator on a quadruped robot enhances its performance compared to a robot without one in tasks such as rapid turning, aerial reorientation, and balancing. The results indicate that the manipulator can improve the agility and stability of the quadruped robot. We believe that our study broadens the roles of manipulators, enabling robotics engineers to utilize them for additional functionalities when integrated with quadruped robots.

However, our research focuses solely on simulations, not real-world applications. Therefore, it remains to be proven that the manipulator can indeed improve the performance of quadruped robots in real-world scenarios. Additionally, we addressed only a few applications in which quadruped robots can be enhanced by using a manipulator as a tail. There are numerous other potential applications that future studies could explore.

\addtolength{\textheight}{-9.5cm}
\bibliographystyle{IEEEtran}
\bibliography{reference.bib}
\end{document}